\documentclass[letterpaper]{article}
\usepackage{spconf,amsmath,amssymb,graphicx}
\usepackage{hyperref}
\usepackage{float}
\usepackage{amsfonts}
\usepackage{algorithm}
\usepackage{algpseudocode}
\usepackage{enumitem}
\setlist{topsep=0pt, leftmargin=*, itemsep=0pt}

\newcommand{\R}[0]{\mathbb{R}}

\newcommand{\mypar}[1]{{\bf #1.}}

\title{Fast Clifford Neural Layers}
\name{Tianxiang Xia, Max Neuwinger, Lin Xiao}
\address{\{xiatia, mneuwinger, linxiao\}@ethz.ch}

\begin{document}
%
\maketitle

\begin{abstract}
Clifford Neural Layers improve PDE modeling by introducing Clifford Algebra into neural networks. In this project we focus on optimizing the inference of 2/3D Clifford convolutional layers and multivector activation layers for one core CPU performance.

Overall, by testing on a real network block involving Clifford convolutional layers and multivector activation layers, we observe that our implementation is 30\% faster than standard PyTorch implementation\cite{CliffordLayersGithub} in relatively large data + network size ($>$L2 cache).

\end{abstract}

\section{Introduction}\label{sec:intro}

\mypar{Motivation} Clifford Neural Layers have various application fields in modeling dynamic systems eg fluid dynamic simulation, weather forecasting. \cite{Brandstetter2022Clifford} \cite{Ruhe2023GeometricClifford}
However, the existing implementation by Microsoft researchers \cite{CliffordLayersGithub} is limited in computation and memory efficiency.

Among the Clifford neural layers, Clifford convolutional and linear layers, and multivector activation layers are necessary even in building simplistic effective Clifford network blocks. This motivated us to explore their optimizations in inference.

\mypar{Contribution} We present optimized C implementations of these layers, corresponding Python interfaces with \texttt{ctype}, and provide an example integration into realistic Clifford network blocks from \cite{CliffordLayersGithub}.

In one core CPU configuration, we achieved maximally 31 FLOPs/cycle (hard compute bound\footnote{i7‑12700KF Alder Lake AVX2 single precision}) for 2/3D Clifford convolutional layers and up to x50 performance of direct C baseline for multivector activation layers. We also explore the optimizations of 1D Clifford convolutional layers (24 FLOPs/cycle) and linear layers.

As for realistic 2D resp.\ 3D network blocks\footnote{cf \texttt{CliffordBasicBlock} in \cite{CliffordLayersGithub}}, we observe a 30\% resp.\ 7\% speedup in relatively large data + network size ($>$L2 cache) setting from standard PyTorch implementation \cite{CliffordLayersGithub} (one core/thread).

The applied and attempted optimizations include avoiding data duplication (eg, Clifford kernel construction), memory layout optimization, (auto)vectorization, (auto)unrolling, prefetching, memory alignment, code style optimization, branching elimination, countable/constant loops, etc. We make realistic assumptions in the most optimized C versions: batch sizes are divisible by 8 and all blades are used in activation layers.

\section{Background on the Algorithm/Application}\label{sec:background}

\mypar{Clifford Algebra} A $\mathbf{k}$-dimensional real Clifford algebra \cite{Ruhe2023GeometricClifford} with {\it signature} $g=\left(g_1, \ldots, g_k\right) \in\{ \pm 1,0\}^k$ is defined as a Ring $(\mathcal{C} \ell_g\left(\mathbb{R}^k\right), +, *)$, generated by a basis $\left\{e_1, \ldots, e_k\right\}$ subject to the relations:
$$
e_i^2=g_i \text{ and } e_i e_j=-e_j e_i ~(i \neq j)
$$
The set of all algebraic elements (called {\it multivectors}) is:
$$
\mathcal{C} \ell_g\left(\mathbb{R}^k\right)=\left\{\sum_{I \subseteq\{1, \ldots, k\}} c_I \cdot e_I \mid c_I \in \mathbb{R}, e_I=\prod_{i \in I} e_i\right\} \simeq \mathbb{R}^{2^k}
$$
where the product $e_I$ is taken in increasing order of indices to ensure consistency under anticommutativity, and $e_{\emptyset}:=1$ is the multiplicative identity. The addition is defined element-wise and the multiplication is defined by distributing over addition and simplifying based on the fore-mentioned relations.

As in \cite{CliffordLayersGithub} we consider a signature is valid if at least one of its elements is non-zero.

A multivector can be described by $(c_I)_{I\subset \{1,\ldots,k\}}\in \R^{2^k}$, which are called the {\it blades} of $x$.  

\mypar{Notations} Throughout the report, we denote $B$=batch size, $C_\mathrm{in}$=number of in-channels, $C_\mathrm{out}$=number of out-channels, $d_\mathrm{filter}$=side length of filter, $d_\mathrm{image}$=side length of image, $d_\mathrm{out}=d_\mathrm{image}-d_\mathrm{filter}+1$=side length of filtered image, $N_B$=number of blades.

\mypar{Clifford convolutional Layers} We consider Clifford convolutional layers with no padding, unit stride, no dilation and equal side length filters and images, ie, conventional convolutional layers with multivector as elements following Clifford Algebra $(+, *)$ cf pseudo-code \ref{alg:clifford_conv}.

\mypar{Clifford linear layers} We consider conventional linear layers with multivector as elements following Clifford Algebra $(+, *)$ cf pseudo-code \ref{alg:clifford_linear}.

\mypar{Multivector activation layers} They are gated functions where a scalar, computed from a subset of input blades, uniformly scales the entire multivector. It supports three aggregation modes (\texttt{agg\_mode}) for computing this scalar: a weighted sum (\texttt{Linear}), a simple sum (\texttt{Sum}), or an average (\texttt{Mean}) of the selected blades, cf pseudo-code \ref{alg:multivector_act}.

\begin{algorithm}
\caption{$k$-D Clifford Convolution}
\label{alg:clifford_conv}
\textbf{Input:}
input $\in \mathcal{C} \ell_g(\mathbb{R}^k)^{B \times C_{\mathrm{in}} \times d_{\mathrm{image}}^k}$,\\
filters $\in \mathcal{C} \ell_g(\mathbb{R}^k)^{C_{\mathrm{in}} \times C_{\mathrm{out}} \times d_{\mathrm{filter}}^k}$,
bias $\in \mathcal{C} \ell_g(\mathbb{R}^k)^{C_{\mathrm{out}}}$\\
\textbf{Output:} $\mathcal{C} \ell_g(\mathbb{R}^k)^{B \times C_{\mathrm{out}} \times d_{\mathrm{out}}^k}$

\begin{algorithmic}[1]
\For{$b = 1$ to $B$}
  \For{$c_{\mathrm{out}} = 1$ to $C_{\mathrm{out}}$}
    \ForAll{$p \in \{0, \dots, d_{\mathrm{image}} - d_{\mathrm{filter}}\}^k$}
      \State $o = 0$
      \For{$c_{\mathrm{in}} = 1$ to $C_{\mathrm{in}}$}
        \ForAll{$q \in \{0, \dots, d_{\mathrm{filter}} - 1\}^k$}
          \State $o = o + \mathrm{filters}[c_{\mathrm{in}}, c_{\mathrm{out}}, q]$
          \State \hspace{2em} $* \ \mathrm{input}[b, c_{\mathrm{in}}, p+q]$
        \EndFor
      \EndFor
      \State $\mathrm{output}[b, c_{\mathrm{out}}, p] = o + \mathrm{bias}[c_{\mathrm{out}}]$
    \EndFor
  \EndFor
\EndFor
\end{algorithmic}
\end{algorithm}

\begin{algorithm}
\caption{$k$-D Clifford Linear Layer}
\label{alg:clifford_linear}
\textbf{Input:}
input $\in \mathcal{C} \ell_g(\mathbb{R}^k)^{B \times C_{\mathrm{in}}}$,
weight $\in \mathcal{C} \ell_g(\mathbb{R}^k)^{C_{\mathrm{in}} \times C_{\mathrm{out}}}$, \\
bias $\in \mathcal{C} \ell_g(\mathbb{R}^k)^{C_{\mathrm{out}}}$ \textbf{Output:} $\mathcal{C} \ell_g(\mathbb{R}^k)^{B \times C_{\mathrm{out}}}$

\begin{algorithmic}[1]
\For{$b = 1$ to $B$}
  \For{$c_{\mathrm{out}} = 1$ to $C_{\mathrm{out}}$}
    \State $o = 0$
    \For{$c_{\mathrm{in}} = 1$ to $C_{\mathrm{in}}$}
      \State $o = o + \mathrm{weight}[c_{\mathrm{in}}, c_{\mathrm{out}}] * \mathrm{input}[b, c_{\mathrm{in}}]$
    \EndFor
    \State $\mathrm{output}[b, c_{\mathrm{out}}] = o + \mathrm{bias}[c_{\mathrm{out}}]$
  \EndFor
\EndFor
\end{algorithmic}
\end{algorithm}

\begin{algorithm}
\caption{Multivector Activation Layer (Baseline Logic)}
\label{alg:multivector_act}
\textbf{Input:} 
input $\in \mathbb{R}^{B \times C \times N_B}$, 
kernel\_indices $\subset \{1, \dots, N_B\}$ of size $K$ \textbf{Optional (Linear):} 
weight $\in \mathbb{R}^{C \times K}$, 
bias $\in \mathbb{R}^{C}$\\
\textbf{Output:} $\mathbb{R}^{B \times C \times N_B}$

\begin{algorithmic}[1]
\For{$b = 1$ to $B$}
  \For{$c = 1$ to $C$}
    \For{$j = 1$ to $N_B$} \Comment{Loop over output blades}
        \State $s = 0$
        \If{agg\_mode is Linear}
            \For{$k = 1$ to $K$}
                \State $s$ += $\mathrm{input}[b,c,\text{kernel\_indices}[k]] * \mathrm{weight}[c,k]$
            \EndFor
            \State $s = s + \mathrm{bias}[c]$
        \ElsIf{agg\_mode is Sum or Mean}
            \For{$k = 1$ to $K$}
                \State $s = s + \mathrm{input}[b,c,\text{kernel\_indices}[k]]$
            \EndFor
            \If{agg\_mode is Mean} $s = s/K$ \EndIf
        \EndIf
        \State $\mathrm{output}[b, c, j] = \mathrm{input}[b, c, j] * \mathrm{sigmoid}(s)$
    \EndFor
  \EndFor
\EndFor
\end{algorithmic}
\end{algorithm}

\section{Optimization Performed}

\mypar{Clifford convolutional layers}
The baseline and all the optimizations follow the same input and output protocol: 
$\mathrm{input}.\mathrm{shape}$=($B$, $C_\mathrm{in}$, $d_\mathrm{image}^k$, $N_\mathrm{B}$), $\mathrm{filters}.\mathrm{shape}$=($N_\mathrm{B}$, $C_\mathrm{in}$, $C_\mathrm{out}$, $d_\mathrm{filter}^k$), $\mathrm{bias}.\mathrm{shape}$=($N_\mathrm{B}$, $C_\mathrm{out})$, $\mathrm{output}.\mathrm{shape}$=($B$, $C_\mathrm{out}$, $d_\mathrm{out}^k$, $N_\mathrm{B}$) where $k$ is both the dimension of the images and Clifford Algebra in consideration, in coherence with \cite{CliffordLayersGithub}.

Our \textbf{Baseline} is a direct C implementation of Clifford 1/2/3D convolutional layers of \cite{CliffordLayersGithub}: it builds a kernel of shape $(C_\mathrm{out}\cdot N_B, C_\mathrm{in}\cdot N_B, d_\mathrm{filter}^k)$\footnote{To build the kernel, it computes  for each blade pair, the contribution coefficient of all filter blades from the input blade to the output blade. See  \texttt{cliffordlayers/cliffordlayers/cliffordkernels.py} in \cite{CliffordLayersGithub} for detailed implementation}, rearrange the input to shape $(B, C_\mathrm{in}\cdot N_B, d_\mathrm{image}^k)$ then apply real algebra convolutional computation by these two. Finally, it rearranges the output from $(B, C_\mathrm{out}\cdot N_B, d_\mathrm{out}^k)$ to $(B, C_\mathrm{out}, d_\mathrm{out}^k, N_B)$.

We notice that, building the kernel leads to $\times N_B$ duplication of data, the memory layouts and computation order do not release the high operation intensity of Clifford Algebra. Our following optimizations tackle main these points.

We first design our optimal data layout and implement a scalar intermediate version in \textbf{Opt1}. In this version, for $k$-D Clifford algebra, there is one hyper-parameter $L$ (referred to as \textit{package length}). We rearrange input data to $(C_\mathrm{in}, d_\mathrm{image}^k, B/L, N_B, L)$, meaning that given in-channel and image position, we group the multivectors in a batch by $L$ and store them in a $(N_B, L)$ package tensor by first indexing the blades. Likewise, the output tensor the rearranged into $(C_\mathrm{out}, d_\mathrm{out}^k, B/L, N_B, L)$ tensor. The filters tensor is rearranged by putting blades of the same multivector together in a filter.

Then in computation, in order, we loop over out-channels, in-channels, input image positions, then packages. We load the whole package into local variables, then loop through filter positions. Given the input image position and filter position, the output position is known. We do an intensive Clifford algebra computation (referred to as \textit{packed computation}) between the package and blades of the filter element and write the results into the output tensor. Finally we rearrange the output tensor back to the protocol format.

We implement this version only for 2D Clifford algebra. This version already achieves $\times 3.33$ speed up from baseline. Its code structure is friendly for vectorization and unrolling, leading to our second optimized version.

\begin{figure}
    \centering
    \includegraphics[width=\linewidth]{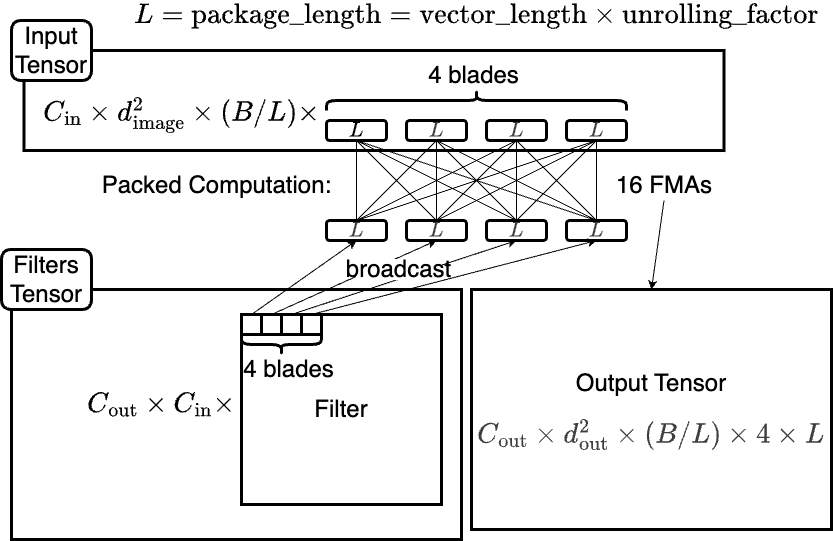}
    \caption{Memory layout and computation graph in optimized Clifford 2D convolutional layers}
    \label{fig:conv_opt}
\end{figure}

In \textbf{Opt2}, to increase ILP, we decompose $L$, the package length from Opt1, into $\text{vector\_length} \times \text{unrolling\_factor}$. The idea is to make packed computation a big chunk of vectorized calculation with high operation intensity. In our case, AVX2 single precision, $\text{vector\_length}=8$. To reduce total FLOP count, we notice that signatures are elements from $\{\pm 1, 0\}$, thus have only finite number of combinations given dimension. So arithmetic expressions can be simplified into only FMADDs/FNMADDs given fixed signature. See Fig.\ \ref{fig:conv_opt} for visual memory layout and computation diagram in 2D case.

We implement a Python script\footnote{\texttt{clib/convolutional/conv\_gen.py} in the repository} to generate C code:
\begin{itemize}
    \item vectorized into AVX2 instructions\footnote{for development reasons, we support also autovectorization in NEON architecture, demonstrating the flexibility of our optimization}
    \item unrolled given unrolling factor
    \item having different functions with simplified arithmetic expressions for different combinations of signatures
    \item supporting 1/2/3D Clifford convolutional layers
    \item in proper code style (packed load, compute, store; scalar replacement; etc.)
\end{itemize}
The ``local operation intensity" is very high in this version. At each vectorized packed computation, there is only read/write from $N_B$ scalars (filter), and $N_B\times\text{unrolling\_factor}$ vectors (output), but there are $N_B^2\times\text{unrolling\_factor}$ vectorized FMAs. (Data from input is read before looping through filter positions.) So the local operation intensity is roughly ${8N_B \over 8 + {1\over \text{unrolling\_factor}}}$. Unrolling increases this local operation intensity and mitigates the performance drawback caused by read-after-write dependencies.

In Opt2, we have optimized to hit the compute bound of our CPU core (32 FLOPs/cycle) for Clifford 2/3D convolutional layers. 
But for the 1D case, there is still room for improvement (25 FLOPs/cycle), mainly because the local operation intensity is not so high as 2/3D cases. We tried to exploit compiler optimization by fixing batch size and thus having a constant sized loop, and tried to use aligned memory operations\footnote{\texttt{clib/convolutional/conv\_gen\_bis.py} in the repository}. These improve the 1D performance only by a small margin.

\mypar{Multivector activation layers}
Our \textbf{Baseline} is a direct C translation of the original PyTorch implementation. It was highly inefficient due to two primary bottlenecks: first, the activation factor was recomputed within the innermost loop over the output blades ($N_B$), causing the expensive \texttt{sigmoid} function \footnote{expf call approximated with 30 cycles} to be called redundantly $N_B$ times per channel. Second, kernel blades were accessed via an indirection array, resulting in a non-sequential gather operation with poor data locality.

Our initial version, \textbf{Opt1}, addressed the redundant computation by hoisting the activation calculation out of the innermost loop. We also applied loop unrolling with dual accumulators to the reduction kernel to increase instruction-level parallelism. To solve the locality issue and enable effective vectorization, \textbf{Opt2} introduced a data layout transformation. In this step, the required kernel blades are gathered from the main tensor into a dense, contiguous \texttt{v\_pack} tensor, which ensures unit-stride memory access in the subsequent computational kernel.

With the data prepared, \textbf{Opt3} introduced SIMD vectorization using AVX2. We processed channels in blocks of eight, using vector instructions for the K-loop reduction (dot product or sum) and a vectorized \texttt{sigmoid256\_ps} function for the non-linearity, which relies on the Intel SVML for an efficient \texttt{\_mm256\_exp\_ps} call. This version remained general, supporting cases where only a subset of blades is used for activation ($K < N_B$) via unaligned loads.

For peak performance, we then specialized the implementation. Assuming the default use case where all blades contribute to the activation ($K=N_B=4$ in 2D and $K=N_B=8$ in 3D) and the number of channels is a multiple of 8. This assumption, implemented in \textbf{Opt4}, allowed us to create dedicated, branch-free code paths for each aggregation mode and blade count, and to use aligned memory accesses. Our final version, \textbf{Opt5}, maximized Instruction-Level Parallelism (ILP) by explicitly unrolling the loop over the block of eight channels and by writing the code in Static Single Assignment (SSA) code style with packed loads, compute and stores. We exposed a large number of independent instructions to the CPU's out-of-order execution engine, allowing it to hide latencies and achieve maximal throughput.

Since the data movement is still the bottleneck, we tried manual prefetching. But this optimization is deleted by icx compiler according to its report.

\mypar{Clifford Linear Layers} The \textbf{Baseline} implementation was a direct C implementation of the Clifford linear layers in \cite{CliffordLayersGithub}. The Clifford Algebra is implemented by building explicitly a kernel of size $(C_\mathrm{out}\cdot N_B, C_\mathrm{in}\cdot N_B)$ from original weight $(N_B, C_\mathrm{out}, C_\mathrm{in})$ with $\times N_B$ data redundancy.

In \textbf{Opt1} we eliminate data rearrangement and kernel build completely, we do the same operations as in baseline but retrieve the data from traced back memory position.

In \textbf{Opt2} we observe that, since the value space of signatures is finite, the FLOP count can be reduced by simplifying arithmetic operations for each signature combination. We write a Python script that generate C functions for each combination with simplified operations. Branching is only performed at the beginning of the program, dependent of the signature combination. 

To improve spatial locality and enable vectorization, we rearrange the input and output arrays in \textbf{Opt3}. Input array is rearranged into $(N_B, C_\mathrm{in})$, the weight is rearranged into $(N_B, C_\mathrm{out}, C_\mathrm{in})$. For each blade pair, we compute a general matrix multiplication (GEMM) and write the result into the output array $(N_B, C_\mathrm{out})$ according to Clifford Algebra. The C code is written by a Python script\footnote{\texttt{clib/matrix\_multiplication/affine\_gen.py} in the repository} that: simplifies operation by writing a different function for each signature combination; vectorizes the GEMM in AVX2. \footnote{Further optimization was not performed, due to it being optimizations on classic GEMM and can be directly done with BLAS\cite{NetlibBLAS}}






\section{Experimental Results}\label{sec:exp}

\mypar{Experimental setup}
All benchmarks were conducted on a 12th Gen Intel\textregistered{} Core\texttrademark{} i7-12700KF processor with 32\,GB of DDR4-3200 memory. The architecture supports AVX2 and SSE2, and its Performance-cores feature a 48\,KB L1 data cache, 1\,MB L2 cache, and a shared 25\,MB L3 cache\cite{PerfDB12700KF}\cite{Intel12700KFSpec}. The read bandwidths of L1, L2, L3 caches and RAM are resp.\ 96\,B/cycle, 64\,B/cycle, 32\,B/cycle, 16\,B/cycle\cite{ChipsCheeseGoldenCove}\cite{WikiGoldenCove}.

To ensure stable and reproducible measurements, we executed all tests on a single P-core with its sibling SMT thread disabled, its frequency locked at a constant 3.6\,GHz via P-state configuration, and Address Space Layout Randomization (ASLR) turned off. All computations used single-precision floating-point numbers as the original Microsoft repository\cite{CliffordLayersGithub}. \texttt{icx/gcc} specs are in plot/caption.



\begin{figure}[h!]
\centering
\includegraphics[width=\linewidth]{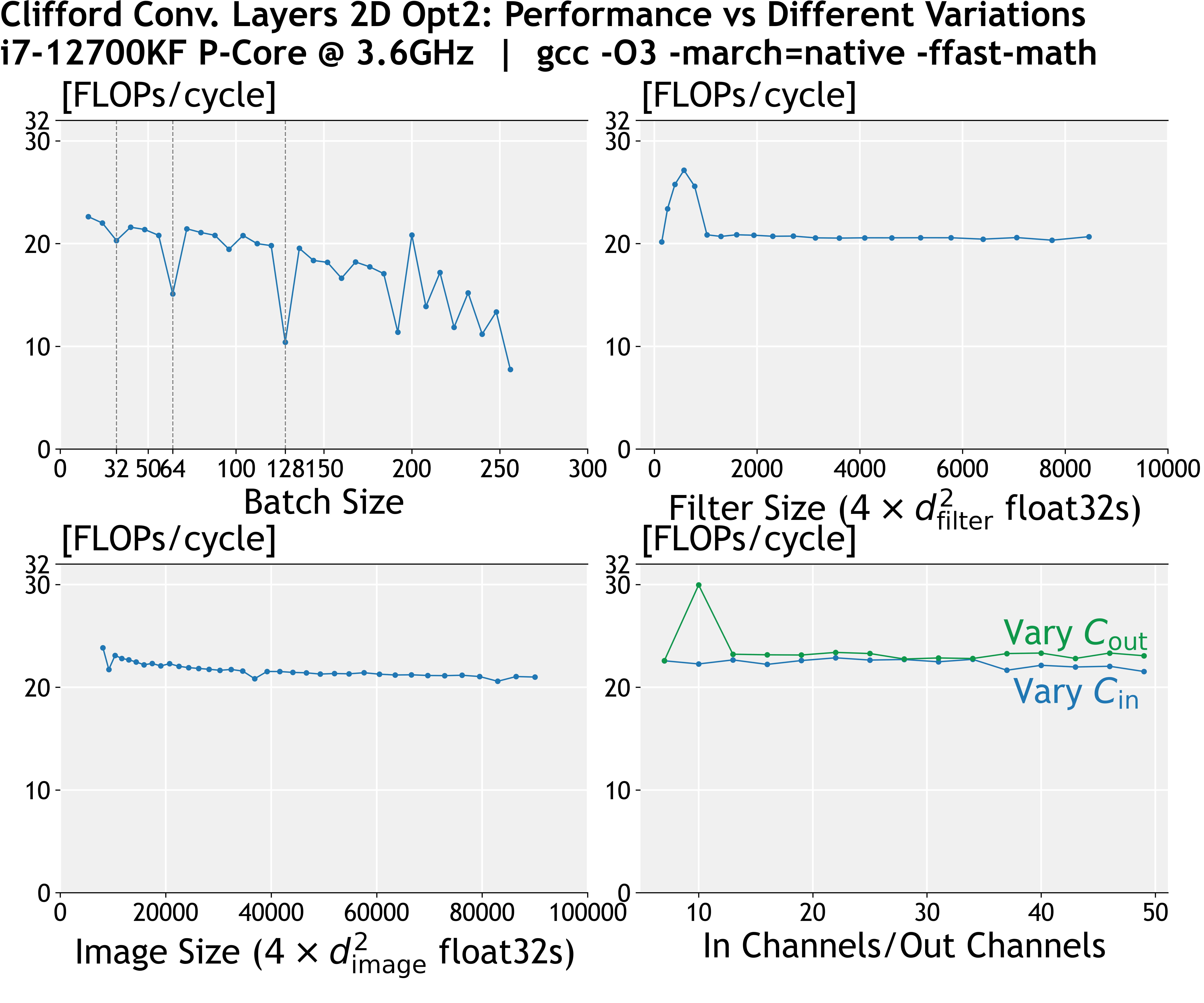}
\caption{Convolutional Layers 2D: Performance vs Different Parameter Variations}
\label{fig:conv_2d_variations}
\end{figure}

\begin{figure}[h!]
    \centering
    \includegraphics[width=\linewidth]{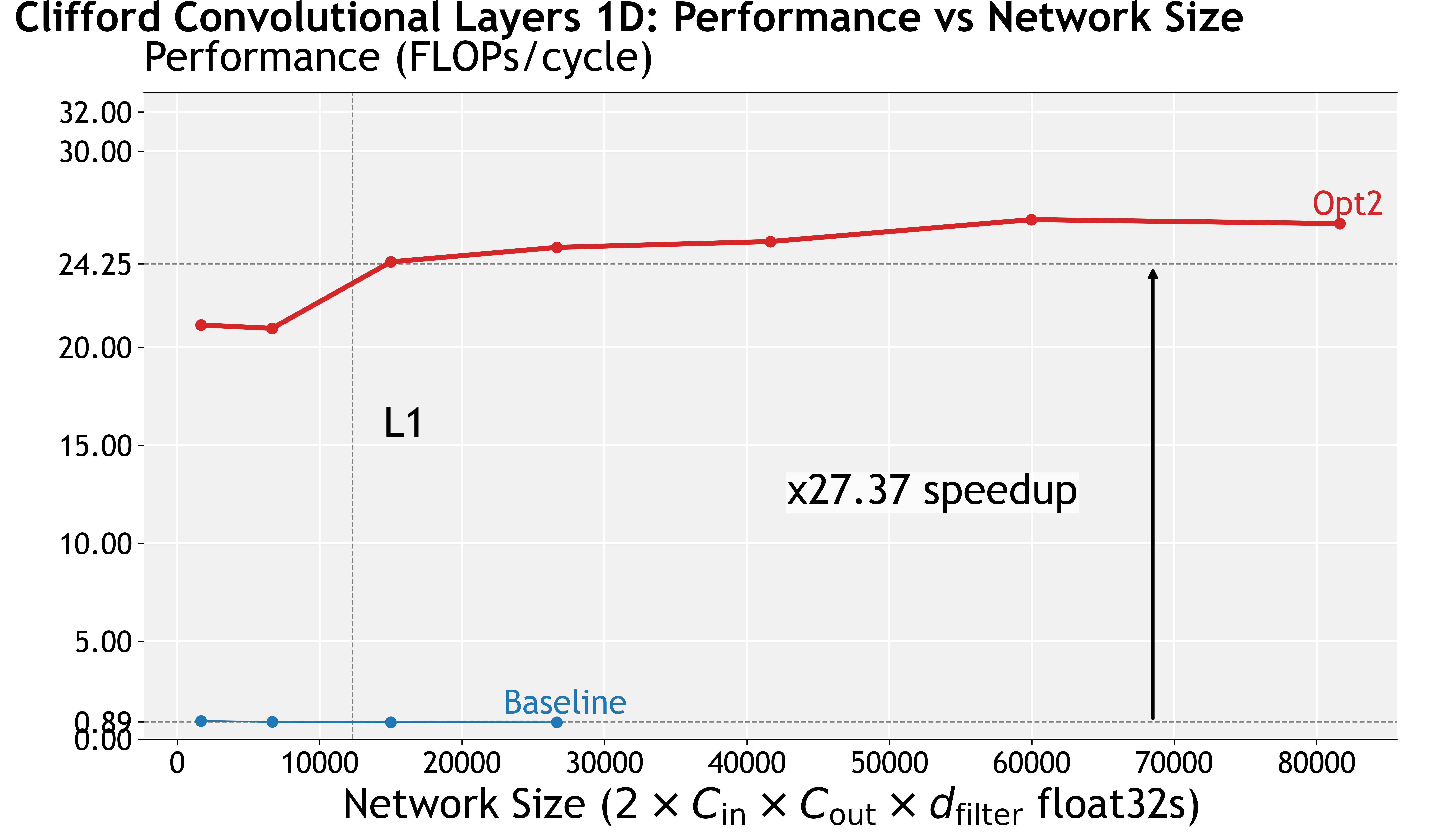}
    \includegraphics[width=\linewidth]{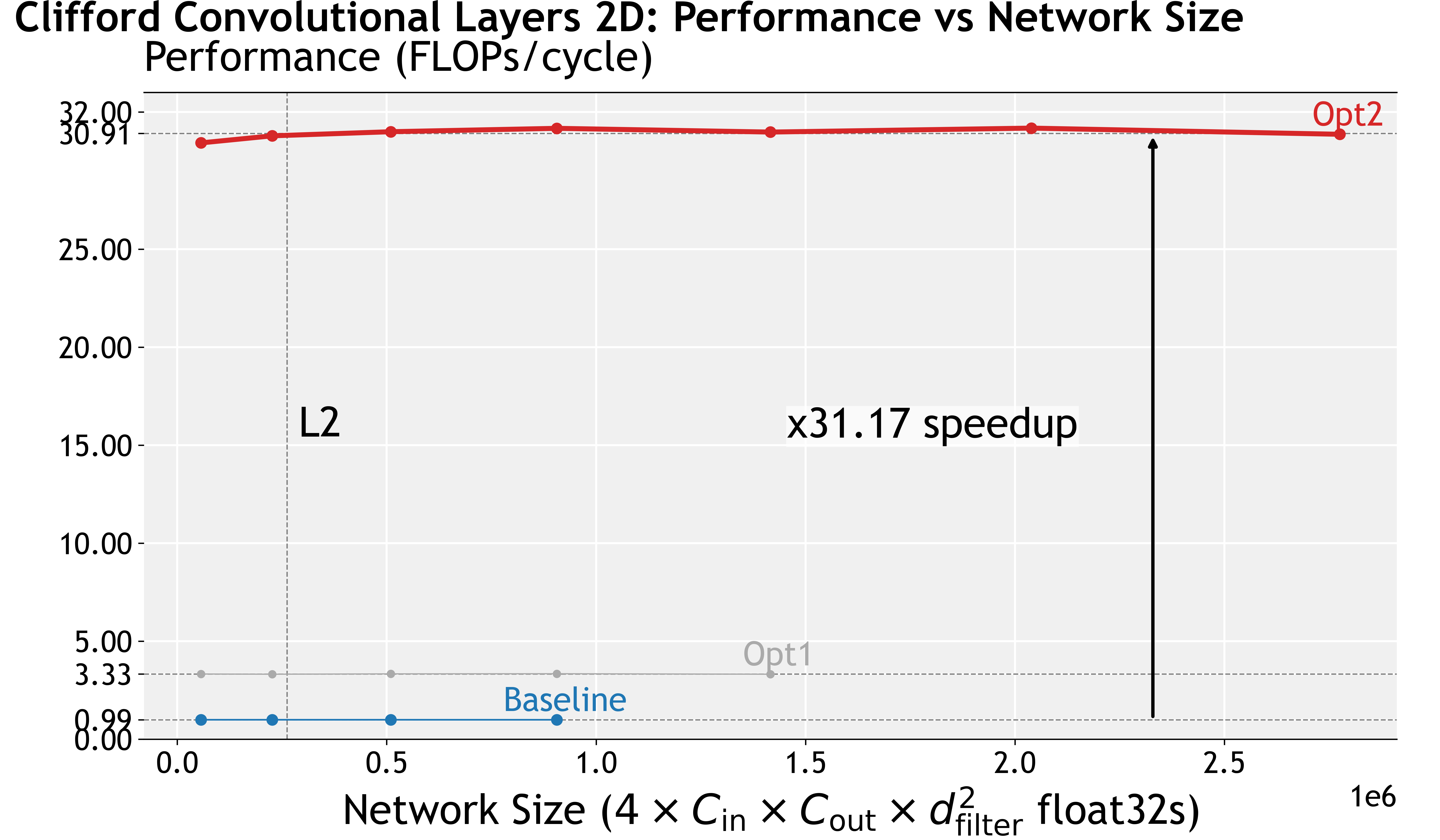}
    \includegraphics[width=\linewidth]{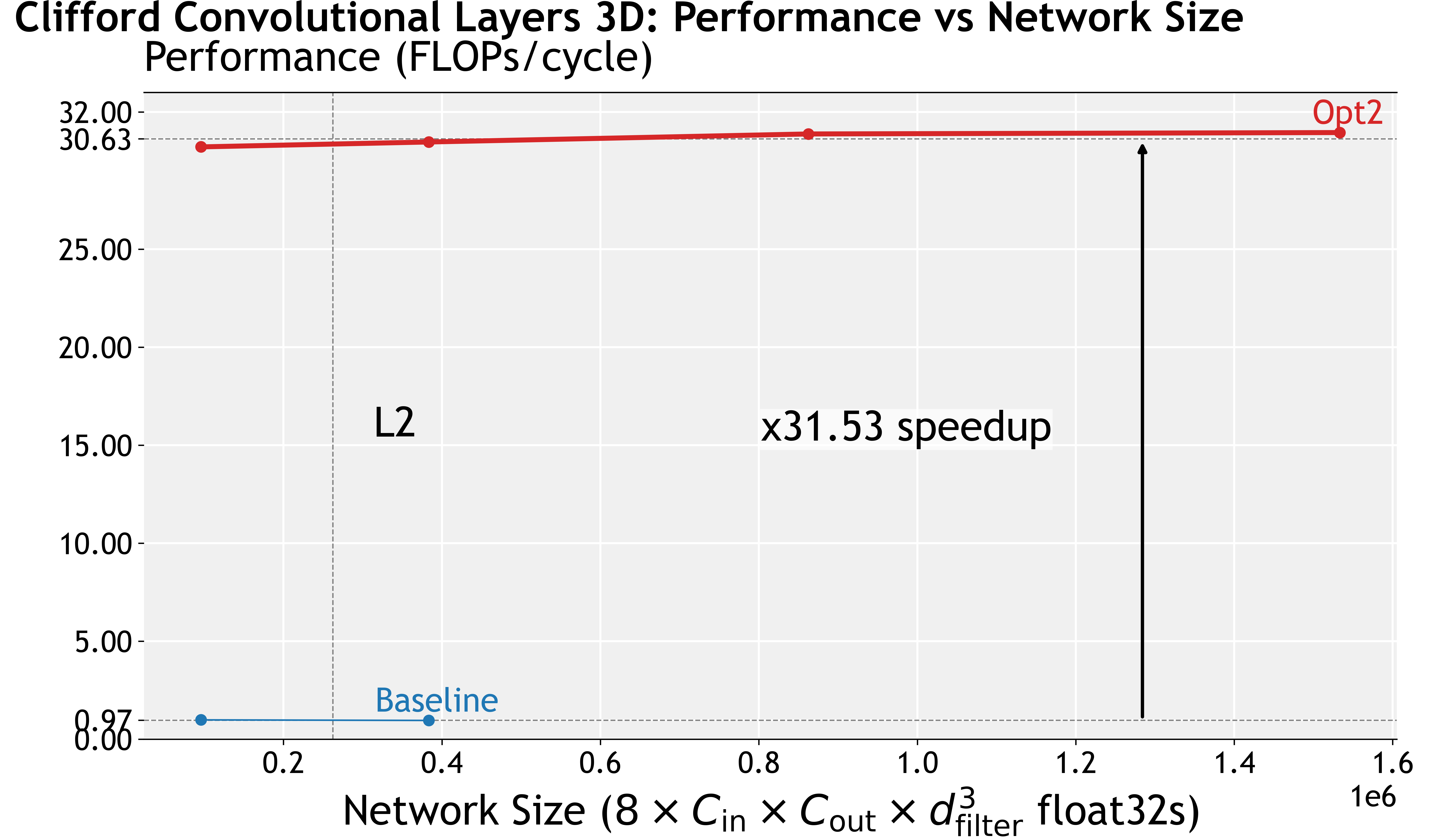}
    \caption{Clifford convolutional Layers: Performance vs Network Size; we fix $(B, d_\mathrm{image}, d_\mathrm{filter})$ and vary $C_\mathrm{in}=C_\mathrm{out}$ to increase the network size: for 2/3D, $(B, d_\mathrm{image}, d_\mathrm{filter})=(8, 60, 17)/(8, 27, 11)$ without unrolling; for 1D, $B=8\times$ (optimal unrolling factor from benchmark for each specific network configuration), $(d_\mathrm{image}, d_\mathrm{filter})=(60, 17)$. All compiled by \texttt{gcc -O3 -march=native -ffast-math}}
    \label{fig:conv_performance}
\end{figure}

\mypar{Clifford convolutional layers} We first analyze the effects of different parameter variations on the performance. As shown in Fig.\ \ref{fig:conv_2d_variations}, the main influence comes from the variation of batch size. Indeed, we rearrange the batch dimension to lower dimensions, which may cause cache conflicts due to 2 power stride when writing into output array. The general performance in the plot is not top performant as $\sim$31 FLOPs/cycle in Fig.\ \ref{fig:conv_performance}, because in order to vary in a wide range one parameter, we have to keep other parameters small, then the overhead of memory rearrangement becomes influential. The reason for the peak in Filter Size and Channels plot remains unclear to us - it may be for the cache friendliness in the particular configurations.

\begin{figure}
\centering
\includegraphics[width=\linewidth]{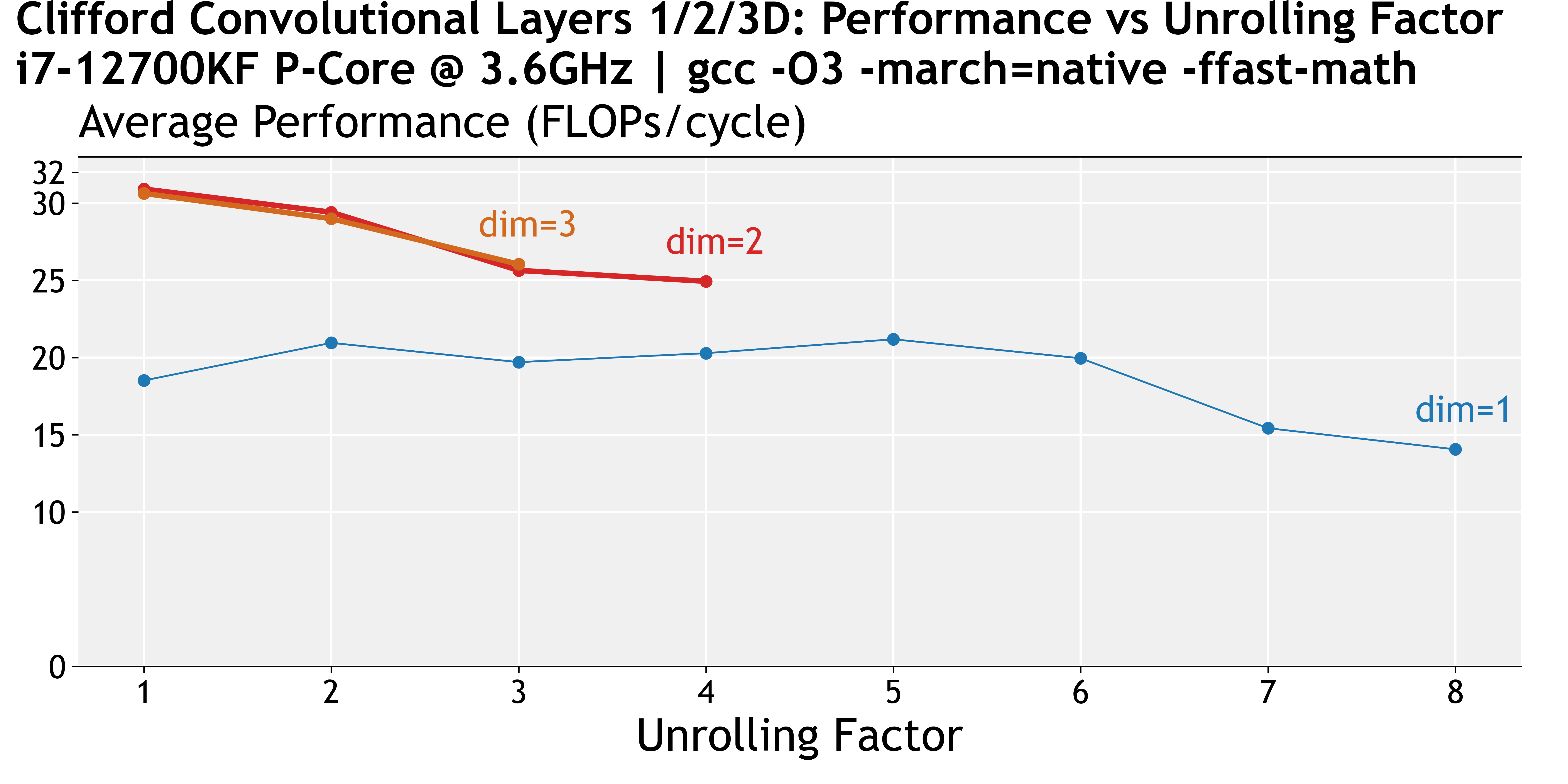}
\caption{Convolutional Layers 3D: Performance vs Unrolling Factor; performance is averaged across benchmark networks in \ref{fig:conv_performance}}
\label{fig:conv_unroll_performance}
\end{figure}

As shown in Fig.\ \ref{fig:conv_unroll_performance}, unrolling actually does not help much in performance (decreases the performance for 2/3D). This is because our CPU is very super-scalar. We tested the same benchmark on a less super-scalar CPU (Intel i5-1038NG7), and unrolling improved significantly the performance for 1/2D, but it never reached compute bound 32 FLOPs/cycle as in 2/3D case on our i7 CPU.

Given the previous results, for the most optimized 2/3D versions, we fix batch size at 8 and do not unroll. Realistically, the user can simply pass multiple 8-sized batches to compute for large batch size.

Finally we test different versions on layers with all parameters comparatively large to eliminate the effect of memory rearrangement. As we can observe in Fig.\ \ref{fig:conv_performance}, our implementations have consistant high performance even when the network size grows bigger than L1 or L2 cache. The performance of 2/3D Clifford convolutional layers hits the 32 FLOPs/cycle compute bound of our CPU.

\begin{figure}[h!]
    \centering
    \includegraphics[width=\linewidth]{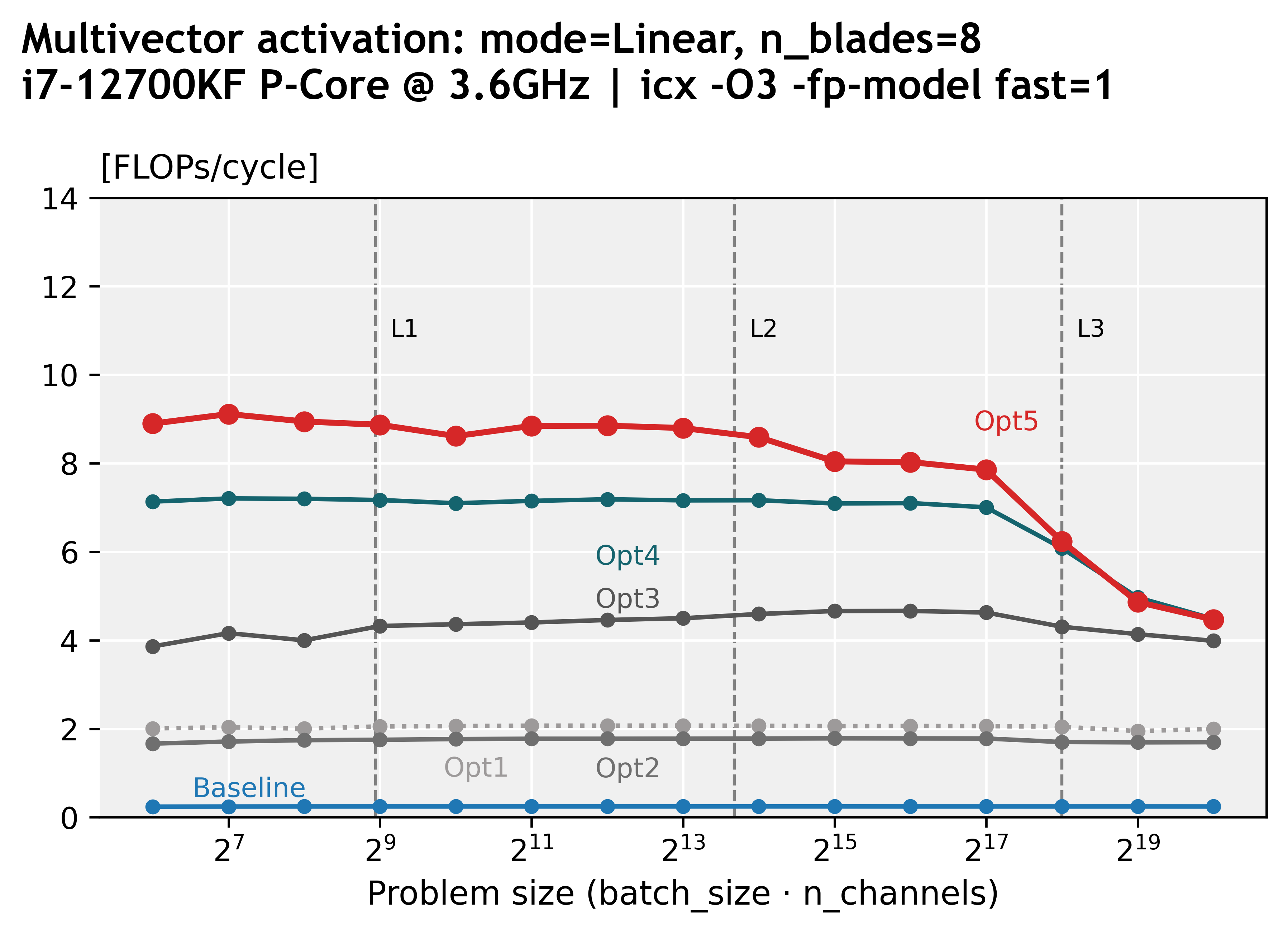}
    \includegraphics[width=\linewidth]{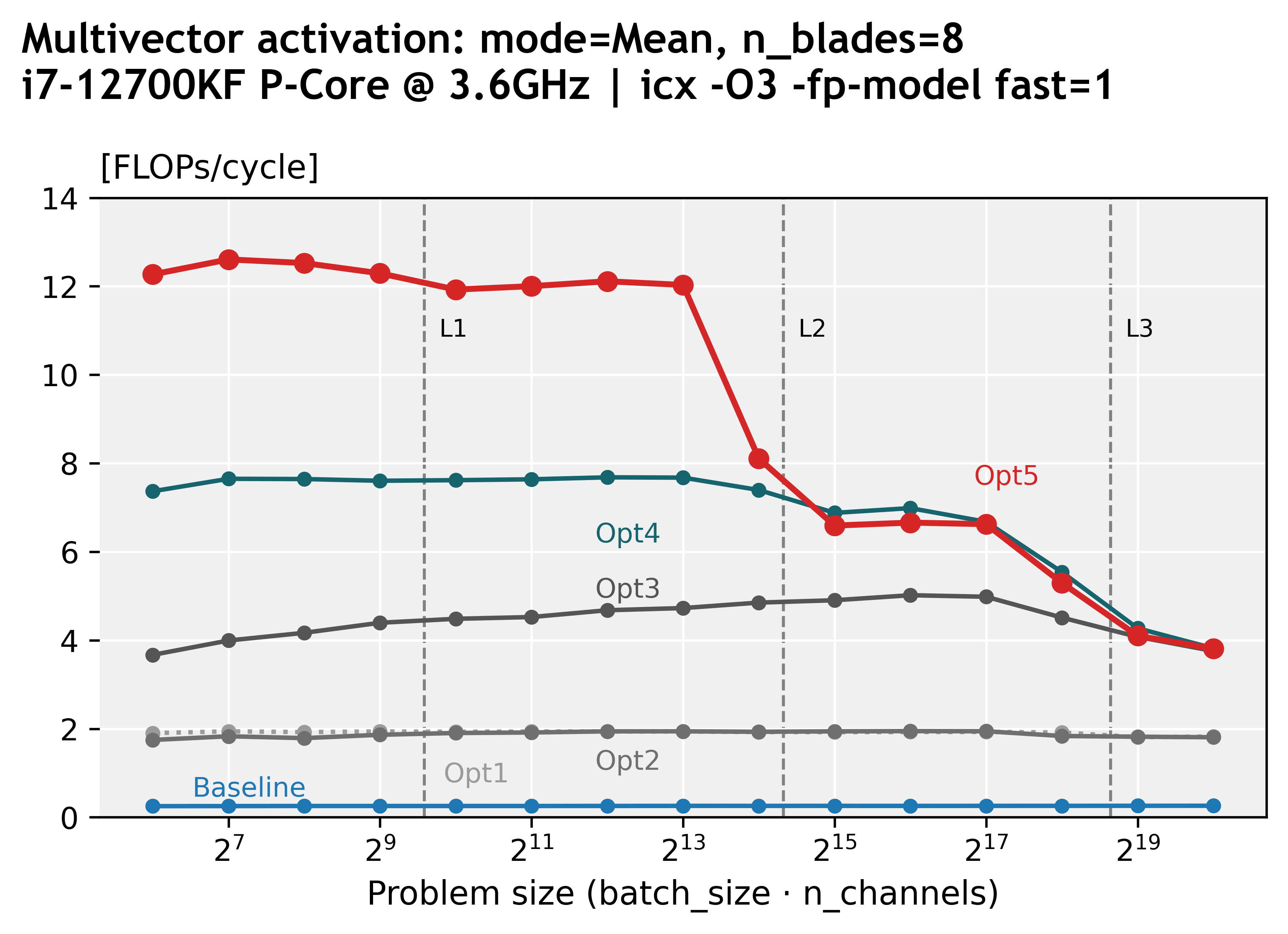}
    \includegraphics[width=\linewidth]{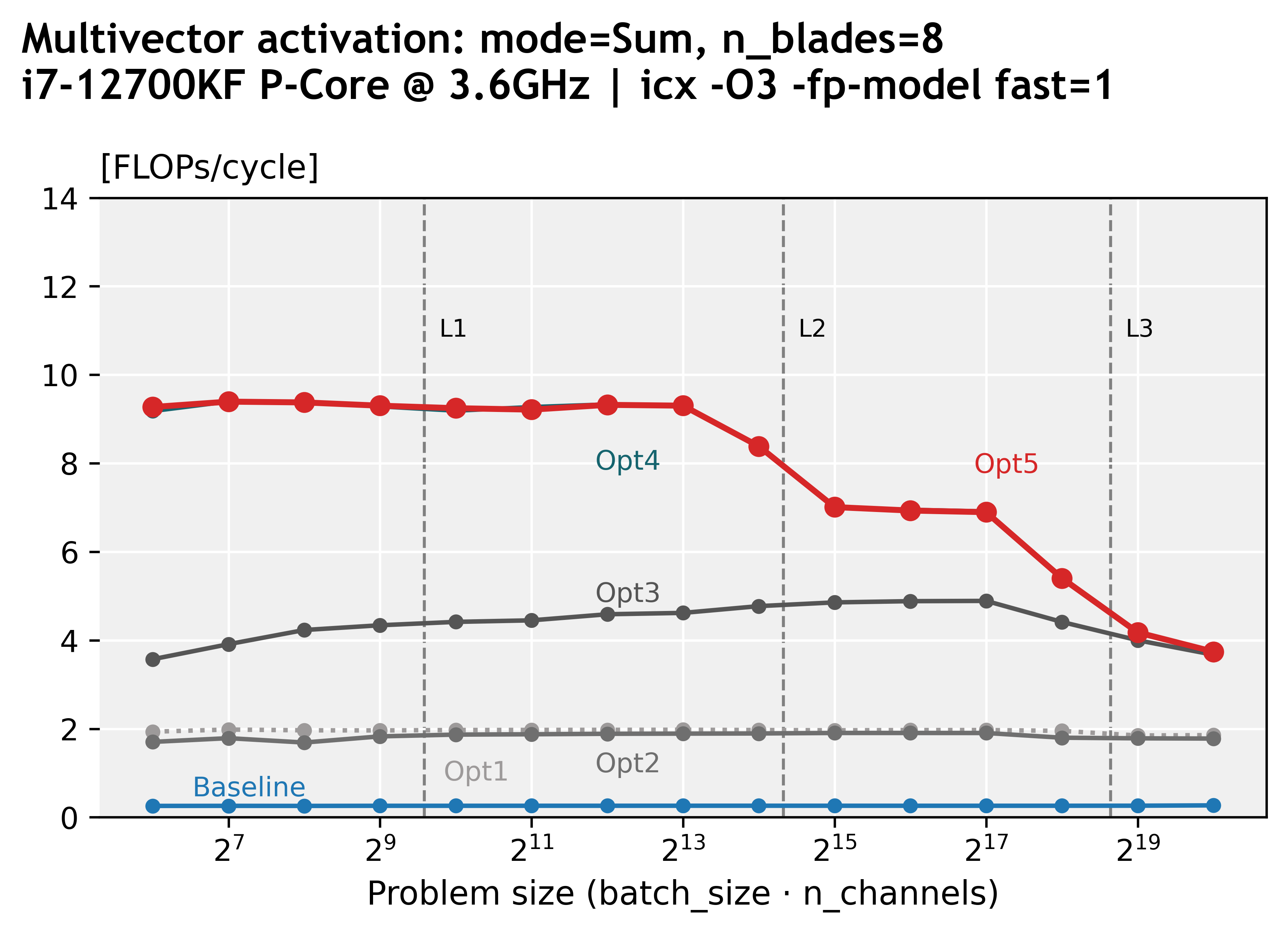}
    \caption{Performance for multivector activation with K=8 blades. The final speedup of Opt5 over the baseline reached up to \textbf{38x} (Linear), \textbf{36.5x} (Sum), and \textbf{50.6x} (Mean).}
    \label{fig:multivec_k8}
\end{figure}

\begin{figure}[h!]
    \centering
    \includegraphics[width=\linewidth]{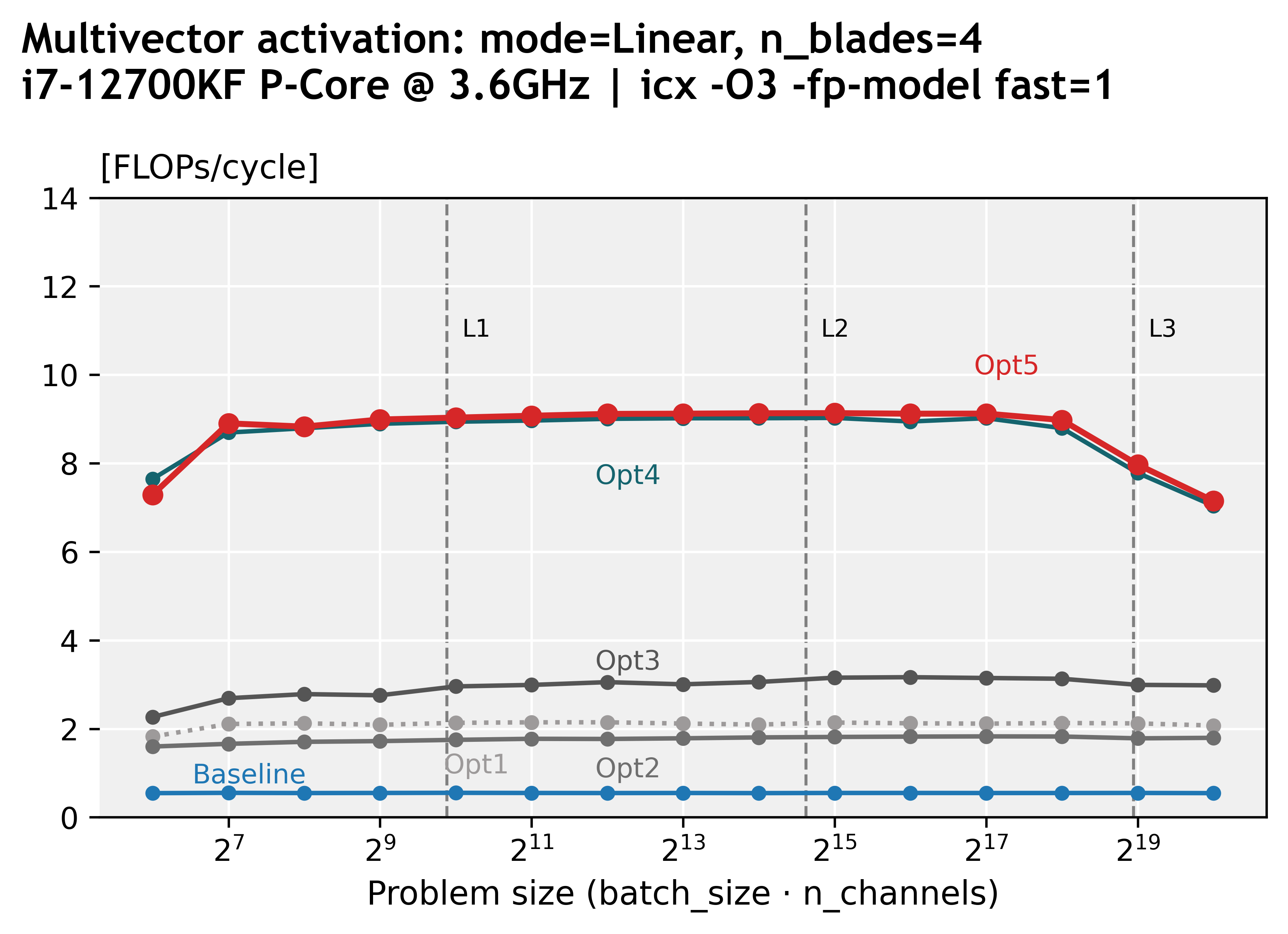}
    \includegraphics[width=\linewidth]{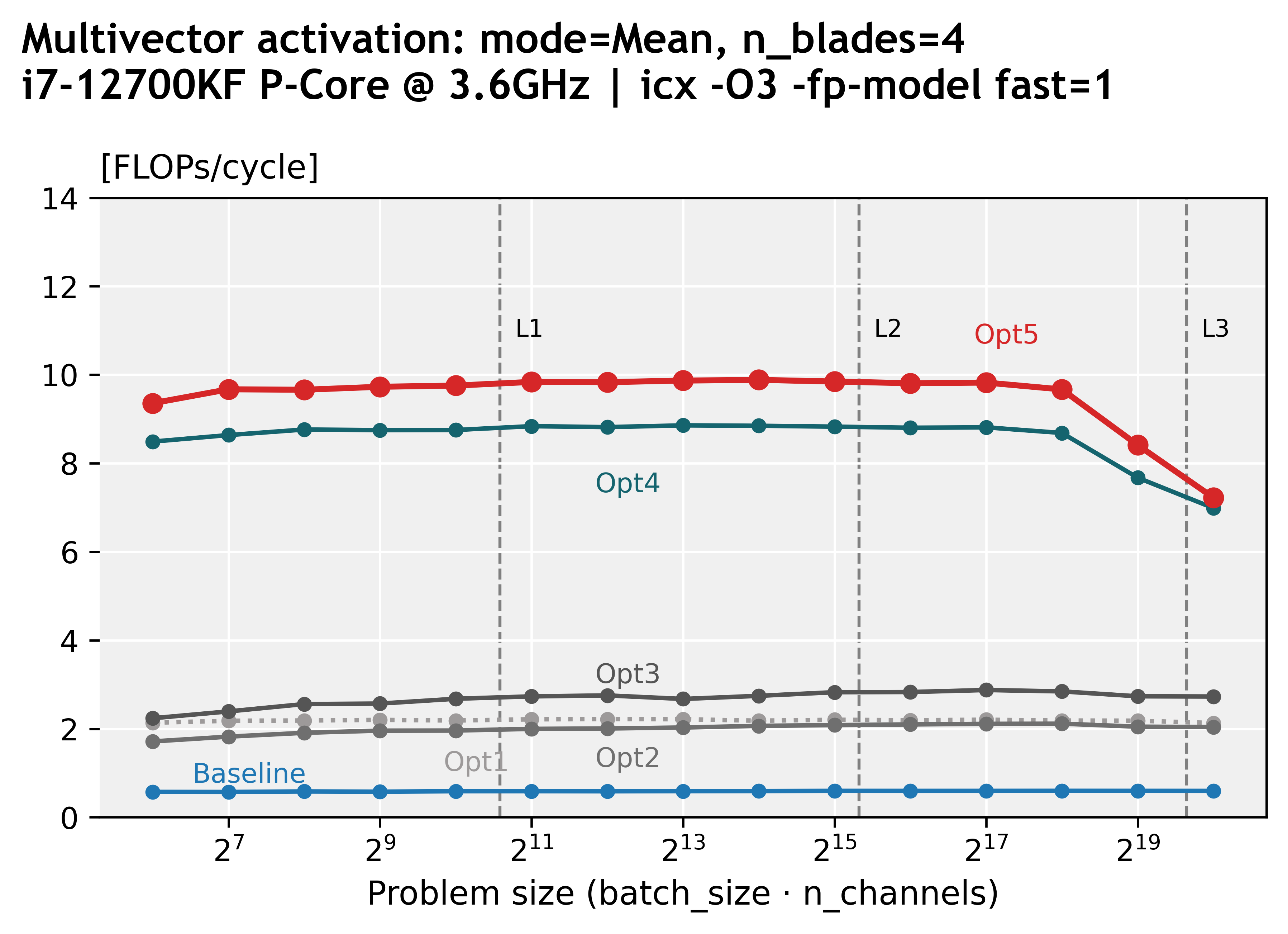}
    \includegraphics[width=\linewidth]{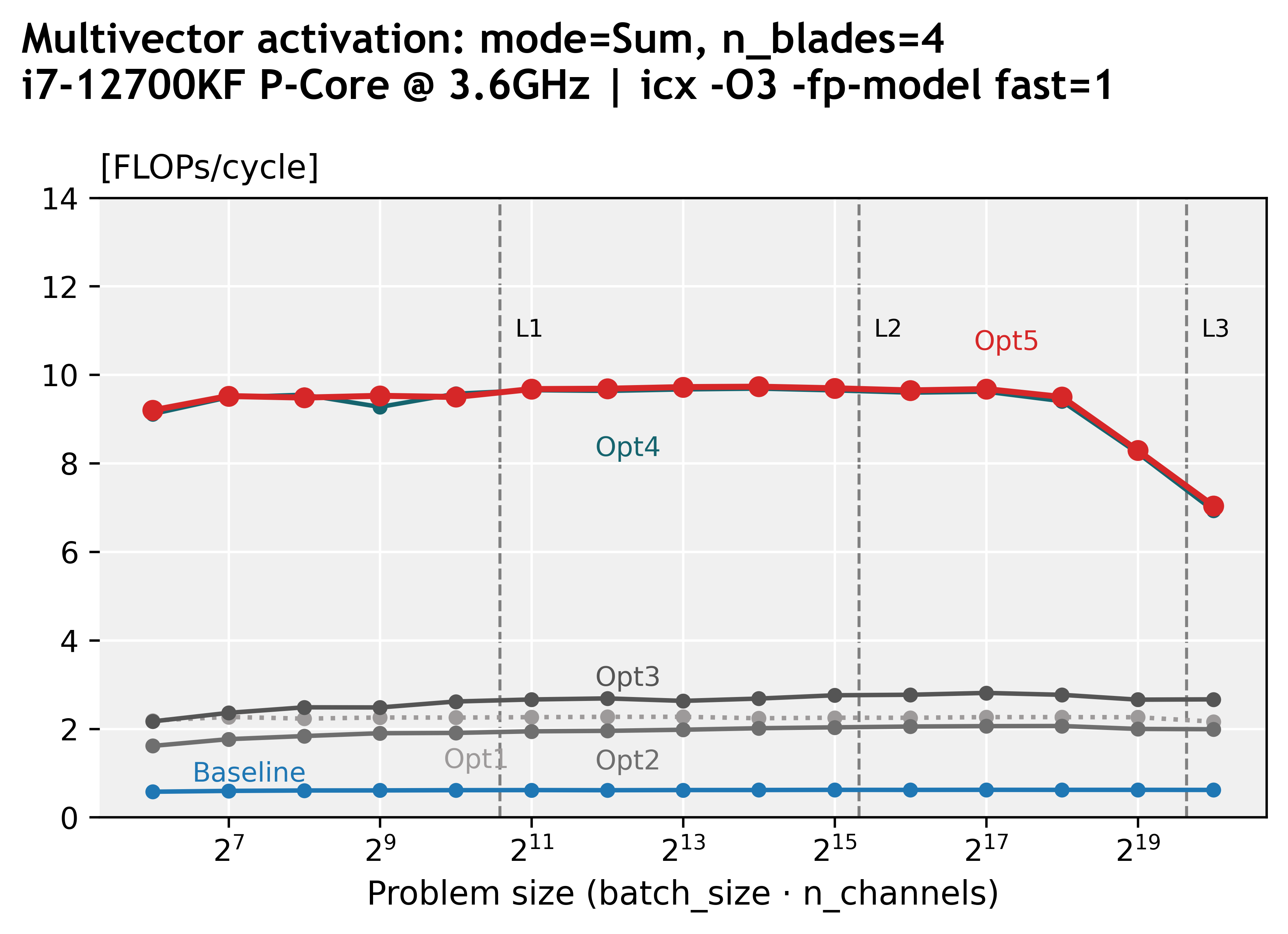}
    \caption{Performance for multivector activation with K=4 blades. The final speedup of Opt5 over the baseline reached up to \textbf{17.0x} (Linear), \textbf{16.2x} (Sum), and \textbf{17.3x} (Mean).}
    \label{fig:multivec_k4}
\end{figure}

\mypar{Multivector activation layers results}
Fig.\ \ref{fig:multivec_k8} ($K=8$) and Fig.\ \ref{fig:multivec_k4} ($K=4$) show the performance gains of our optimizations. The plots depict FLOPs/cycle vs.\ problem size ($B \times C$, i.e., batch size $\times$ channels) across all three aggregation modes, with a clear progression explained by microarchitectural profiling and compiler analysis.

Our analysis begins with the \textbf{Baseline}, which exhibits very low, constant performance. \textbf{Opt1} achieves a significant speedup by eliminating redundant computations. Interestingly, \textbf{Opt2} performs nearly as well as Opt1, despite being algorithmically simpler (no unrolling or dual accumulators). This demonstrates the impact of the packed data layout, which provides sequential memory access. Opt3 provides the expected SIMD speedup. The specialized versions, \textbf{Opt4} and \textbf{Opt5}, show more nuanced behavior. For both $K=8$ and $K=4$, Opt5 shows a clear advantage in the Linear and Mean modes, where its fully unrolled structure allows the CPU to hide instruction latencies. In the Sum modes, however, the performance of Opt4 and Opt5 is nearly identical, suggesting the compiler already generated near-optimal code.

Cache effects are prominent in the $K=8$ case. A performance drop occurs once the input data exceeds the L2/3 cache size, most severely in the memory-sensitive Sum and Mean modes. A key observation is in the Mean mode: Opt4's performance is stable across cache boundaries, whereas Opt5's performance drops sharply. This suggests that while Opt5's high-ILP design is exceptionally effective when data is in-cache, its burst of memory requests saturates the memory pipeline when stalls occur.

A microarchitectural analysis using Intel VTune Profiler reveals the underlying reasons for these shifts. Our initial versions (Baseline, Opt1, Opt2) were entirely scalar, with VTune reporting 0\% FP Vector utilization. Their performance was severely limited by a `Core Bound' bottleneck, with scalar `Divider Stalls' accounting for 17-22\% of execution time. With \textbf{Opt3}, the introduction of SIMD instructions (16\% FP Vector uOps) caused `Divider Stalls' to plummet to 5\%. As we progress to \textbf{Opt4} and \textbf{Opt5} (35-42\% FP Vector uOps), the primary bottleneck shifts dramatically to the back-end. VTune shows `Back-End Bound' stalls increasing from 20\% in Opt1 to over 70\% in our best versions, a combination of `Memory Bound' stalls (from $<5\%$ to $>20\%$) and `Core Bound' stalls due to contention for vector execution ports. The VTune data perfectly explains the Opt5-Mean anomaly: it registers the highest `Back-End Bound' of all runs (78.6\%), with a massive 12.3\% `DRAM Bound' stall time, confirming its memory access pattern stalled the pipeline.

Compiler optimization reports from `icx' provide a final layer of evidence. The \textbf{Baseline}'s poor performance is explained by severe register pressure (`58 spills`) and loop multiversioning due to pointer aliasing uncertainty. Our manual unrolling in \textbf{Opt1} exacerbated this issue, increasing pressure to `79 spills'. In contrast, \textbf{Opt2}'s cleaner structure enabled the compiler to perform `loop collapsing' and reduce spills, explaining its performance. As we moved to SIMD, \textbf{Opt4}'s two-phase structure drastically cut register pressure to just `14 spills'. Critically, the compiler report for \textbf{Opt5} shows that its full, SSA-style unrolling did not increase register pressure over Opt4. This confirms that our final optimization provided a ``free" boost in ILP by exposing more independent instructions to the CPU's out-of-order core without incurring the cost of additional register spills.


\begin{figure} [h!]
    \centering
    \includegraphics[width=1.05\linewidth]{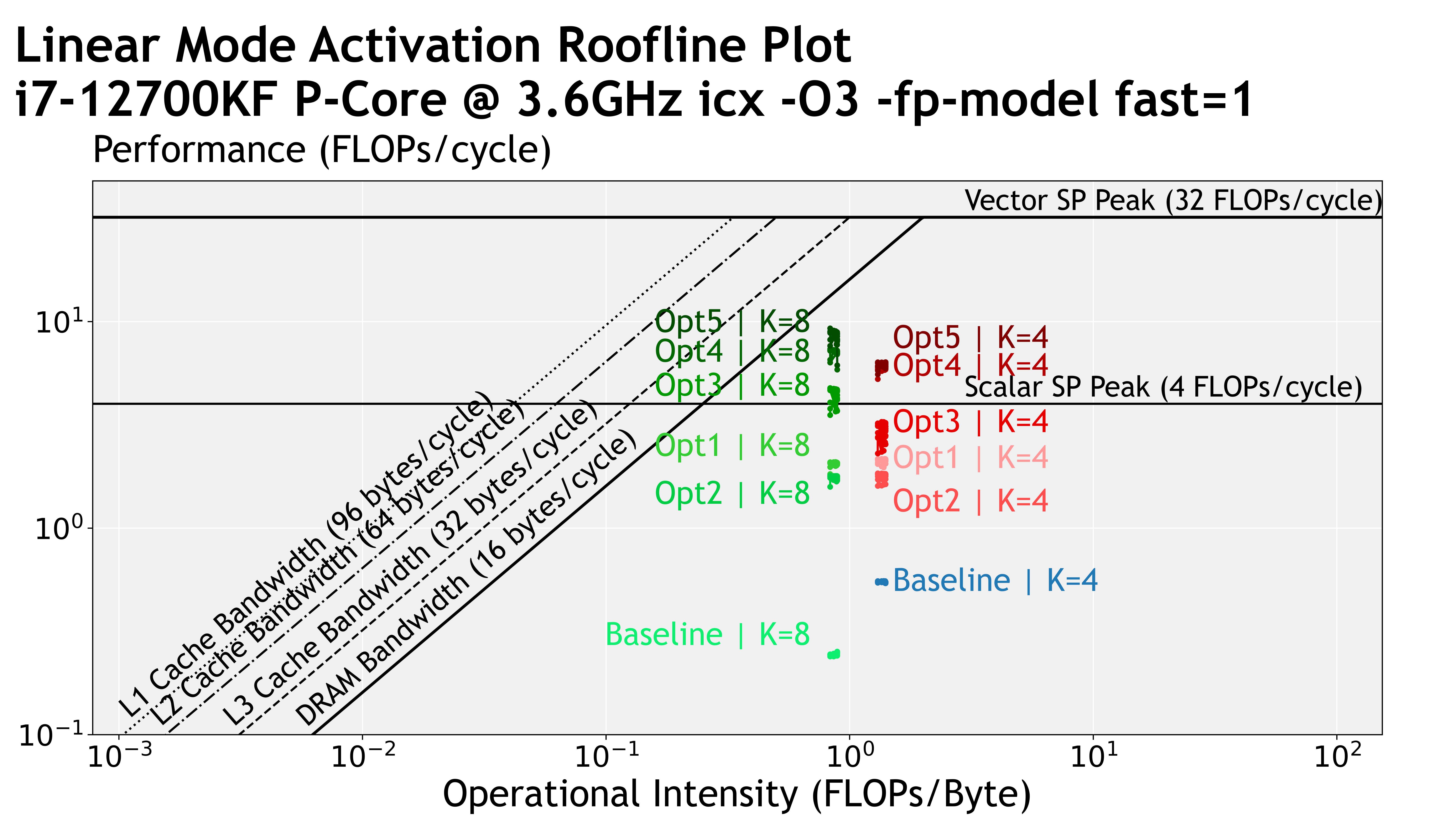}
    \caption{Roofline plot for multivector activation layer (Linear mode shown; Sum and Mean modes yielded similar results); Operational intensity was estimated manually, considering only compulsory cache misses.}
    \label{fig:multivec_roofline}
\end{figure}

From the roofline plot \ref{fig:multivec_roofline}, all scalar optimizations are compute-bound, while vectorized versions are more memory-bound. K=4 is more compute-bound with a smaller working set than K=8, explaining the better cache behavior.

\begin{figure} [h!]
\centering
\includegraphics[width=\linewidth]{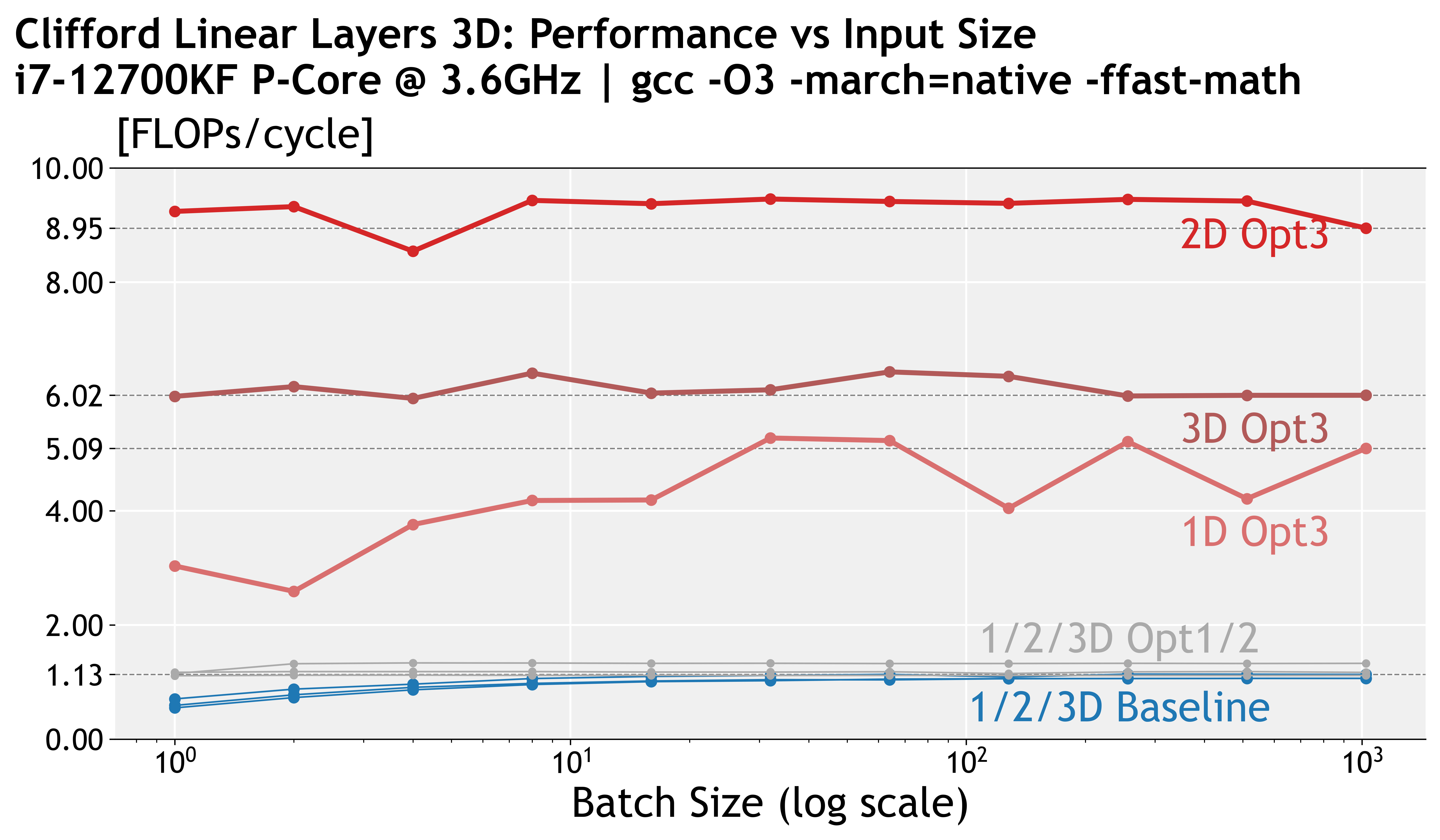}
\caption{Linear Layers: Performance vs Input Size: $C_\mathrm{in}=C_\mathrm{out}=32$; flops counts are from most optimized version}
\label{fig:linear_performance}
\end{figure}

\mypar{Clifford linear layers results} We have achieved in average a 4.49x/8.40x/5.66x speedup for 1/2/3D Clifford linear layers cf Fig.\ \ref{fig:linear_performance}. Opt1 and Opt3 are the most effective. Opt2 indeed significantly reduced the FLOP count, but those operations could be highly parallelized with other operations and thus did not provide an evident performance boost (cf Table \ref{tab:linear_performance_metrics} VTune results).

\begin{table}
\small
\centering
\begin{tabular}{|l|c|c|c|c|}
\hline
\textbf{Metric} & \textbf{Baseline} & \textbf{Opt1} & \textbf{Opt2} & \textbf{Opt3} \\
\hline
Clockticks & 20.34B & 15.81B & 15.78B & 1.89B \\
Instructions Retired & 76.83B & 59.49B & 54.49B & 4.55B \\
3+ Ports Utilized & 80.2\% & 91.2\% & 88.2\% & 60.8\% \\
\hline
\end{tabular}
\caption{Performance metrics across Clifford linear layer optimizations on a 2D network with 128 batches and 32 input/output channels.}
\label{tab:linear_performance_metrics}
\end{table}

\mypar{Comparison with PyTorch} Finally we conduct a comparison with realistic networks\footnote{cf \texttt{benchmarking/with\_pytorch/benchmark.py} in repository for a script building those networks adapted from \texttt{CliffordBasicBlock2d} in \cite{CliffordLayersGithub} and benchmarking runtime on relative large scale (input data + network size $>$ L2 cache)} utilizing both 2/3D Clifford convolutional layers and mutlivector activation layers cf Table\ref{tab:backend_times}. As activation layers are not the bottleneck, our C implementations improve only the performance in 2D by a small amount. Our most optimized convolutional layers always improve performance, the speed up is around ${3051 \over 2340} - 1\simeq 30\%$ in 2D and ${3479\over 3238}-1\simeq 7\%$ in 3D.

\begin{table}[h!]
\centering
\begin{tabular}{|c|c|c|c|}
\hline
\textbf{Dim} & \textbf{Impl.\ Act} & \textbf{Impl.\ Conv} & \textbf{Avg.\ Time (ms)} \\
\hline
2D & py/c & py & 3051/3042 \\
2D & py/c & c  & 2340/2316 \\
3D & py/c & py & 3479/3487 \\
3D & py/c & c  & 3238/3270 \\
\hline
\end{tabular}
\caption{Average execution time for different layer implementations: `c' denotes our most optimized C version employed by \texttt{ctype} in Python; `py' denotes direct PyTorch implementation\cite{CliffordLayersGithub}.}
\label{tab:backend_times}
\end{table}

\section{Conclusions}
This project presents comprehensive optimization, performance analysis and profiling of Clifford neural layers targeting our specific i7-12700KF CPU. It demonstrates how to release the high operation intensity of Clifford convolutional layers and achieve one core maximum performance (2/3D), how to reduce a Clifford linear layer to multiple GEMM computations with minimal FLOP count and no data duplication, and how to optimize the multivector activation layer into a AVX2-vectorized, high-throughput kernel with up to \(50\times\) speed-up. In addition to the applied and effective optimizations, various other optimizations are suggested, tried and analysed as they may be valuable in other settings.





\newpage

\section{Contributions of Team Members (Mandatory)}



\mypar{Tianxiang Xia} Designed and implemented baseline and optimizations for Clifford linear layers and Clifford convolutional layers. Plotted for those layers. Wrote script for comparison with PyTorch.

\mypar{Max Neuwinger} Designed and implemented baseline and optimizations for multivector activation layers. Wrote benchmarking and plotting scripts for them, and executed all benchmarks and VTune profiling on the target CPU.

\mypar{Lin Xiao} Analyzed results. Wrote benchmarking code for linear and convolution layer and Vtune script. Did cost analysis and bottleneck identification with Vtune analysis for all layers. Plotted Roofline plots of multivector activation layers.

\bibliographystyle{IEEEbib}
\bibliography{bibl_conf}

\end{document}